\documentclass[width=10in]{article}

\usepackage{microtype}
\usepackage{graphicx}
\usepackage{subfigure}
\usepackage{booktabs} % for professional tables

\usepackage{hyperref}

\usepackage{amsmath}
\usepackage{amssymb}
\usepackage{mathtools}
\usepackage{amsthm}
\usepackage{natbib}

\usepackage{algorithm}
\usepackage{algorithmic}

\usepackage{geometry}
\usepackage[capitalize,noabbrev]{cleveref}

\theoremstyle{plain}
\newtheorem{theorem}{Theorem}[section]

\newtheorem{corollary}[theorem]{Corollary}
\theoremstyle{definition}

\theoremstyle{remark}

\usepackage[textsize=tiny]{todonotes}

\title{Actor-Critic Methods using Physics-Informed Neural Networks: Control of a 1D PDE Model for Fluid-Cooled Battery Packs}

\author{Amartya Mukherjee$^1$\thanks{Corresponding author.\hfil\break e-mail: a29mukhe@uwaterlooo.ca}~, Jun Liu$^1$
\\
\small{$^1$Department of Applied Mathematics, University of Waterloo, Waterloo, Ontario N2L 3G1, Canada}}

\date{}

\begin{document}

% \twocolumn[




% \begin{authorlist}
% \end{authorlist}

% \affiliation{to}{Department of Applied Mathematics, University of Waterloo, Waterloo, Ontario, Canada}

% \correspondingauthor{Amartya Mukherjee}{a29mukhe@uwaterloo.ca}


\vskip 0.3in
% ]




\maketitle

\begin{abstract}
This paper proposes an actor-critic algorithm for controlling the temperature of a battery pack using a cooling fluid. This is modeled by a coupled 1D partial differential equation (PDE) with a controlled advection term that determines the speed of the cooling fluid. The Hamilton-Jacobi-Bellman (HJB) equation is a PDE that evaluates the optimality of the value function and determines an optimal controller. We propose an algorithm that treats the value network as a Physics-Informed Neural Network (PINN) to solve for the continuous-time HJB equation rather than a discrete-time Bellman optimality equation, and we derive an optimal controller for the environment that we exploit to achieve optimal control. Our experiments show that a hybrid-policy model that updates the value network using the HJB equation and updates the policy network identically to PPO achieves the best results in the control of this PDE system.\\

\noindent
{\bf keywords:} Actor-Critic Method, Physics-Informed Neural Network, Fluid Cooled Battery Packs, Hamilton Jacobi Bellman Equation
\end{abstract}

\section{Introduction}
In recent years, there has been a growing interest in Reinforcement Learning (RL) for continuous control problems. RL has shown promising results in environments with unknown dynamics through a balance of exploration in the environment and exploitation of the learned policies. Since the advent of REINFORCE with Baseline, the value network in RL algorithms has shown to be useful towards finding optimal policies as a critic network (\cite{SutBar}). This value network continues to be used in state-of-the-art RL algorithms today.

Proximal Policy Optimization (PPO) is an actor-critic method introduced by \cite{PPO2017}. It limits the update of the policy network to a trust region at every iteration. This ensures that the objective function of the policy network is a good approximation of the true objective function and forces smooth and reliable updates to the value network as well.

In discrete-time RL, the value function estimates returns from a given state as a sum of the returns over time steps. This value function is obtained by solving the Bellman Optimality Equation. On the other hand, in continuous-time RL, the value function estimates returns from a given state as an integral over time. This value function is obtained by solving a partial differential equation (PDE) known as the Hamilton-Jacobi-Bellman (HJB) equation \cite{Munos1999_2}. Both equations are difficult to solve analytically and numerically, and therefore the RL agent must explore the environment and make successive estimations.

The introduction of physics-informed neural networks (PINNs) by \cite{RAISSI2019686} has led to significant advancements in scientific machine learning. PINNs leverage auto-differentiation to compute derivatives of neural networks with respect to their inputs and model parameters exactly. This enables the laws of physics (described by ODEs or PDEs) governing the dataset of interest to act as a regularization term for the neural network. As a result, PINNs outperform regular neural networks on such datasets by exploiting the underlying physics of the data.

Control of PDEs is considered to be challenging compared to control of ODEs. Works such as \cite{7509652} introduced the backstepping method for the boundary control of reaction-advection-diffusion equations using kernels. For PDE control problems where the control input is encoded in the PDE, the HJB equation has been used (\cite{SIRIGNANO20181339},\cite{HJBPDE}). Works from control of ODEs have been used by writing the PDE as an infinite-dimensional ODE.

To the best of our knowledge, this paper is the first to explore the intersection between PINNs and RL in a PDE control problem. We discretize the PDE as an ODE to derive an HJB equation. In order to force the convergence of the value network in PPO towards the solution of the HJB equation, we utilize PINNs to encode this PDE and train the value network. Upon deriving the HJB equation, we also derive an optimal controller. We introduce two algorithms: HJB value iteration and Hamilton-Jacobi-Bellman Proximal Policy Optimization (HJBPPO), that train the value function using the HJB equation and use the optimal controller. The HJBPPO algorithm shows superior performance compared to PPO and HJB value iteration on the PackCooling environment.

\section{Preliminaries}

\subsection{The 1D pack cooling problem}

The 1D system for fluid-cooled battery packs was introduced by \cite{Kato_2021} and is modeled by the following coupled PDE:

\begin{align}
	\label{eq:PDE1}
	u_t(x,t)=&-D(x,t)u_{xx}(x,t)+h(x,t,u(x,t))+\frac{1}{R(x,t)}(w-u)
\end{align}
\begin{equation}
	\label{eq:PDE2}
	w_t=-\sigma(t)w_x+\frac{1}{R(x,t)}(u-w),
\end{equation}
with the following boundary conditions:
\begin{equation}
	\label{eq:bc1}
	u_x(0,t)=u_x(1,t)=0
\end{equation}
\begin{equation}
	\label{eq:bc2}
	w(0,t)=U(t)
\end{equation}
where $u(x,t)$ is the heat distribution across the battery pack, $w(x,t)$ is the heat distribution across the cooling fluid, $D(x,t)$ is the thermal diffusion constant across the battery pack, $R(x,t)$ is the heat resistance between the battery pack and the cooling fluid, $h(x,t,u)$ is the internal heat generation in the battery pack, $U(t)$ is the temperature of the cooling fluid at the boundary, and $\sigma(t)$ is the transport speed of the cooling fluid, which will be the controller in this paper.

The objective of the control problem in this paper is to determine $\sigma(t)$ such that $u(x,t)$ is as close to zero as possible. The transport speed $\sigma(t)$ is strictly non-negative so the cooling fluid travels only in the positive $x$-direction. We restrict $\sigma(t)$ to $[0,1]$.

\subsection{Hamilton-Jacobi-Bellman equation}

%%%%%%%% TRANSITION: WHY ODE? %%%%%%%%%%%

To achieve optimal control for the 1D PDE pack cooling problem, we will utilize works from control theory for ODEs. Consider a controlled dynamical system modeled by the following equation:

\begin{equation}
	\dot x = f(x,\sigma), \quad x(t_0)=x_0, 
\end{equation}
where $x(t)$ is the state and $\sigma(t)$ is the control input. In control theory, the optimal value function $V^*(x)$ is useful towards finding a solution to control problems (\cite{Munos1999}):

\begin{align}
	\label{eq:value1}
	V^*(x)=\sup_{\sigma}\frac{1}{\Delta t}\int_{t_0}^{\infty}\gamma^{\frac{t}{\Delta t}} L(&x(\tau;t_0,x_0,\sigma(\cdot)),\sigma(\tau))d\tau,
\end{align}
where $L(x,\sigma)$ is the reward function, $\Delta t$ is the time step size for numerical simulation, and $\gamma$ is the discount factor. The following theorem introduces a criteria for assessing the optimality of the value function (\cite{Liberzon2012}, \cite{RLOFC}).

\begin{theorem}
	\label{thm:1}
	A function $V(x)$ is the optimal value function if and only if:
	\begin{enumerate}
		\item $V\in C^1(\mathbb{R}^n)$ and $V$ satisfies the Hamilton-Jacobi-Bellman (HJB) Equation
		\begin{align}
			\label{eq:HJB1}
			&(\gamma-1)V(x)+\sup_{\sigma\in U}\{L(x,\sigma)+\gamma\Delta t\nabla_xV^T(x)f(x,\sigma)\}=0
		\end{align}
		for all $x\in\mathbb{R}^n$.
		\item For all $x\in\mathbb{R}^n$, there exists a controller $\sigma^*(\cdot)$ such that:
		\begin{align}
			&(\gamma-1)V(x)+L(x,\sigma^*(x))+\gamma\Delta t\nabla_xV^T(x)f(x,\sigma^*(x))
			\nonumber\\&=(\gamma-1)V(x)+\sup_{\hat \sigma\in U}\{L(x,\hat \sigma)+\gamma\Delta t\nabla_xV^T(x)f(x,\hat\sigma)\}.\label{eq:optimalcontrol}
		\end{align}
	\end{enumerate}
\end{theorem}

The proof of part 1 of this theorem is in Appendix \ref{app:pf1}. The HJB equation will be used in this paper to determine a new loss function for the value network $V(x)$ in this pack cooling problem and an optimal controller $\sigma^*(t)$.

\section{Related work}

The HJB equation we intend to solve is a first-order quasi-linear PDE.
The use of HJB equations for continuous RL has sparked interest in recent years among the RL community as well as the control theory community and has led to promising works.
\cite{HJBDQL} introduced an HJB equation for Q Networks and used it to derive a controller that is Lipschitz continuous in time. This algorithm has shown improved performance over Deep Deterministic Policy Gradient (DDPG) in three out of the four tested MuJoCo environments without the need for an actor network. 
\cite{DHJB} introduced a distributional HJB equation to train the FD-WGF Q-Learning algorithm. This models return distributions more accurately compared to Quantile Regression TD (QTD) for a particle-control task. Finite difference methods are used to solve this HJB equation numerically.
Furthermore, the authors mentioned the use of auto-differentiation for increased accuracy of the distributional HJB equation as a potential area for future research in their conclusion.

The use of neural networks to solve the HJB equation has been an area of interest across multiple research projects.
\cite{DBLP:journals/corr/JiangCCT16} uses a structured Recurrent Neural Network to solve the HJB equation and achieve optimal control for the Dubins car problem.
\cite{4267720} uses the Pineda architecture (\cite{pineda1987generalization}) to estimate partial derivatives of the value function with respect to its inputs. They used the iterative least squares method to solve the HJB equation. This algorithm shows convergence in several control problems without the need for an initial stable policy.

RL for PDE control is a challenging field that has been of interest to the machine learning community lately.
\cite{7963427} introduces the Deep Fitted Q Iteration to solve a boundary control problem for a 2D convection-diffusion equation. The model stabilizes the temperature in the environment without encoding any knowledge of the governing PDE.
\cite{SIRIGNANO20181339} develops the DGM algorithm to solve PDEs. They use auto-differentiation to compute first-order derivatives and Monte Carlo methods to estimate higher-order derivatives. This algorithm was used to solve the HJB equation to control a stochastic heat equation and achieved an error of 0.1\%.
\cite{HJBPDE} approximates the solution to the HJB equation using polynomials. This was used to control a semilinear parabolic PDE.

PINNs have been used for the control of dynamical systems in recent works.
\cite{antonelo2021physics} uses a PINN for model predictive control of a dynamical system over a long time interval. The PINN takes the initial condition, the control input, and the spatial and temporal coordinates as input and estimates the trajectory of the dynamical system while repeatedly shifting the time interval towards zero to allow for long-range interval predictions.
\cite{NICODEMUS2022331} uses a PINN-based model predictive control for the tracking problem of a multi-link manipulator.
\cite{djeumou2022neural} uses a PINN to incorporate partial knowledge about a dynamical system such as symmetry and equilibrium points to estimate the trajectory of a controlled dynamical system.

The use of a PINN to solve the HJB equation for the value network was done by \cite{HJBPINN} in an optimal feedback control problem setting. The paper achieves results similar to that of the true optimal control function in high-dimensional problems.

\section{HJB control of the pack cooling problem}

In this section, we will connect the pack cooling PDE model with the HJB equation to derive a new loss function for the value network $V(u,w)$ using the HJB equation and an optimal controller. The HJB equation has been useful in finding optimal controllers for systems modeled by ODEs. In \cite{HJBPDE}, the controlled PDE system has been discretized in space to form an ODE that can be used in the HJB equation. Similarly, to form the HJB equation for this paper, we need to write equations \ref{eq:PDE1} and \ref{eq:PDE2} as an ODE.

\subsection{ODE discretization of PDE}
\label{sec:ODE}

We can write equations \ref{eq:PDE1} and \ref{eq:PDE2} as an ODE by discretizing it in the $x$ variable. By letting $\Delta x=\frac{1}{N_x}$ where $N_x$ is the number of points we choose to discretize the system along the x-axis, we arrive at a $2N_x$ dimensional ODE:

\begin{equation}
	\label{eq:ODE1}
	\dot{\hat U}=-DA\hat U+h(\hat U)+\frac{1}{R}(\hat W-\hat U)
\end{equation}
\begin{equation}
	\label{eq:ODE2}
	\dot{\hat W}=-\sigma(t)B\hat W+\frac{1}{R}(\hat U-\hat W),
\end{equation}
where
$$\hat W(t)=\begin{pmatrix}w(x_1,t)\\\vdots\\w(x_{N_x},t)\end{pmatrix}, \hat U(t)=\begin{pmatrix}u(x_1,t)\\\vdots\\u(x_{N_x},t)\end{pmatrix},
$$
and $A\hat U$ is a second-order discretization of $u_{xx}$, e.g., 
\begin{equation*}
	[A\hat U]_k=\frac{u(x_{k+1},t)-2u(x_k,t)+u(x_{k-1},t)}{\Delta x^2}, 
\end{equation*}
$B\hat W$ is a second-order discretization of $w_{x}$, e.g., 
\begin{equation*}
	[B\hat W]_k=\frac{w(x_{k+1},t)-w(x_{k-1},t)}{2\Delta x}.
\end{equation*}

\subsection{Derivation of the optimal controller}

The ODE system derived in section \ref{sec:ODE} can be used in the HJB equation to determine a loss function and an optimal controller.

\begin{theorem}
	\label{thm:HJB}
	Let $u(\cdot,t),w(\cdot,t)\in L_2[0,1]$. With $\sigma(t)\in[0,1]$ and the reward function $L(U_t,W_t,\sigma_t)=-||U_{t+1}||_2^2\Delta x$, the HJB equation for the 1D pack cooling problem is:
	\begin{align}
		&(\gamma-1)V-||u(\cdot,t+\Delta t)||^2\nonumber\\&+\langle V_u(u(\cdot,t),w(\cdot,t)),u_t(\cdot,t)\rangle\nonumber\\&+\frac{1}{R}\langle V_w(u(\cdot,t),w(\cdot,t)),u(\cdot,t)-w(\cdot,t)\rangle\nonumber\\&+\max(0,-\langle V_w(u(\cdot,t),w(\cdot,t)),w_x(\cdot,t)\rangle)=0
	\end{align}
	where $||\cdot||$ is the $L_2[0,1]$ norm and $\langle\cdot,\cdot\rangle$ is the $L_2[0,1]$ inner product.
\end{theorem}

The proof of this theorem is in Appendix \ref{app:pfThm}. Theorem \ref{thm:1} shows that there exists a controller that satisfies equation \ref{eq:optimalcontrol}. This allows us to determine an optimal controller, as shown in the following corollary:

\begin{corollary}
	\label{cor:OptimalController}
	Let $w(\cdot,t)\in L_2[0,1]$. With $\sigma(t)\in[0,1]$ and the reward function $L(U_t,W_t,\sigma_t)=-||U_{t+1}||_2^2\Delta x$, provided the optimal value function $V^*(u,w)$ with $V^*_w(\cdot,t)\in L_2[0,1]$, the optimal controller for the 1D pack cooling problem is:
	\begin{equation}
		\sigma^*(t)=\begin{cases}1,&\langle V^*_w(u(\cdot,t),w(\cdot,t)),w_x(\cdot,t)\rangle<0,\\0,&\text{otherwise},\end{cases}
	\end{equation}
	where $\langle\cdot,\cdot\rangle$ is the $L_2[0,1]$ inner product.
\end{corollary}

The proof of this corollary is in Appendix \ref{app:pfCor}. These results will be used in our algorithms to achieve optimal control of the pack cooling problem.

\section{Algorithm}

For the control of the PDE, we introduce two algorithms. The first algorithm, called HJB Value Iteration, uses only a value network and exploits the HJB equation and optimal controller derived in Theorem \ref{thm:HJB} and Corollary \ref{cor:OptimalController}. The second algorithm, called HJBPPO, is a hybrid-policy model that uses policy network updates from PPO and value network updates from HJB Value Iteration.

To define these algorithms, we first define two loss functions. The first loss function is derived from the proof of theorem \ref{thm:HJB}.

\begin{align}
	\label{eq:HJBLoss2}
	MSE_f = \frac{1}{T}\sum_{t=0}^{T-1}(&(\gamma-1)V(\hat U_t,\hat W_t)\nonumber\\&-||\hat U_{t+1}||_2^2\Delta x\nonumber\\&+\nabla_UV^T(\hat U_t,\hat W_t)\dot{\hat U}_t\Delta t\nonumber\\&+\frac{1}{R}\nabla_WV^T(\hat U_t-\hat W_t)\Delta t\nonumber\\&+\max(0,-\nabla_WV^TB\hat W))^2\Delta t
\end{align}

The second loss function provides an initial condition. At $u(x,T)=0,w(x,T)=-R(x,t)=-2$, we have: $u(x,T)=0$ and $u_t(x,T)=0$. As a result, we have $L(0,-R(x,t))=0$ and $L_t(0,-R(x,t))=0$. This shows us that $u(x,T)=0,w(x,T)=-R(x,t)$ is considered a stable point that maximizes the reward. 
%Numerically solving this system with $\sigma(t)=0$ shows that the cummulative return after $100$ time steps ($1$ second) is roughly $-0.3426$ with $u(x,T+1)\approx 0.1900$. The authors estimate that this return can be further maximized by using a non-zero controller. 
Thus, we choose to let $V(0,-R(x,t))=0$ be the Dirichlet boundary condition for the HJB equation. This leads to the second loss function:

\begin{equation}
	\label{eq:HJBInitLoss}
	MSE_u=(V(0,-R(x,t)))^2. 
\end{equation}

Since the value function achieves its global maximum at $u(x,T)=0,w(x,T)=-2$, this means the derivatives of $V$ must be zero along all directions. Thus, we choose to let $\frac{\partial V}{\partial n}=0$ at $u(x,T)=0,w(x,T)=-R(x,t)$ along every normal be the Neumann boundary condition for the HJB equation. This leads to the third loss function:

\begin{equation}
	\label{eq:HJBNeumannLoss}
	MSE_n=||\nabla_UV(0,-R(x,t))||_2^2+||\nabla_WV(0,-R(x,t))||_2^2
\end{equation}

We derived an optimal controller in corollary \ref{cor:OptimalController}. Gym environments recommend that actions be in the range $[-1,1]$. We can use the proof of the optimal controller in Appendix \ref{app:pfCor} to derive a way of selecting actions:

\begin{equation}
	\label{eq:OptimalControlExperiment}
	a_t=-\text{sign}(\nabla_WV^TB\hat W(t))
\end{equation}

The algorithms introduced in this paper will focus on minimizing both of the loss functions defined above and using the optimal controller.

\subsection{HJB value iteration}

The HJB Value Iteration trains the loss function without the need for an actor-network. We treat the value network as a PINN, using auto-differentiation to estimate gradient vectors to compute the loss in equation \ref{eq:HJBLoss2} and the control in equation \ref{eq:OptimalControlExperiment}. At every time step, it uses the controller given in equation \ref{eq:OptimalControlExperiment}. It updates the value network using the loss functions as shown above. The method is provided in Algorithm \ref{alg:HJBVal}.

\begin{algorithm}[tb]
	\caption{HJB Value Iteration}
	\label{alg:HJBVal}
	\begin{algorithmic}[1]
		\STATE Initiate value network parameter $\phi$
		\STATE Run the control as given in equation (\ref{eq:OptimalControlExperiment}) in the environment for $T$ timesteps and observe samples $\{(s_t,a_t,R_t,s_{t+1})\}_{t=1}^{T}$.
		\STATE Compute the value network loss as: $J(\phi)=MSE_f+MSE_u+MSE_n$ described in equations (\ref{eq:HJBLoss2}), (\ref{eq:HJBInitLoss}), and (\ref{eq:HJBNeumannLoss})
		\STATE Update $\phi\leftarrow\phi-\alpha_2\nabla_\phi J(\phi)$
		\STATE Run steps 2--4 for multiple iterations
	\end{algorithmic}
\end{algorithm}

\subsection{HJBPPO}

HJBPPO is an algorithm that combines policy optimization from PPO with HJB value iteration. This is implemented by modifying the PPO implementation by \cite{pytorch_minimal_ppo}.

To facilitate exploration of the environment and exploitation of the models, we introduce an action selection method that uses the policy network and equation \ref{eq:OptimalControlExperiment} with equal probability, as shown in Algorithm \ref{alg:action}. Upon running the policy $\pi_\theta$, we sample from a distribution $N(\mu,s)$ where $\mu$ is the output from the policy network. We initiate $s$ to $0.3$ and decrease it by $0.01$ every $1000$ episodes until it reaches $0.1$. After sampling an action from the normal distribution, we clip it between $-1$ and $1$.

\begin{algorithm}[tb]
	\caption{HJBPPO action selection}
	\label{alg:action}
	\begin{algorithmic}[1]
		\STATE Retrieve state $s_t$, policy network parameter $\theta$ and value network parameter $\phi$
		\STATE Sample $i\in\{0,1\}$
		\STATE\textbf{if} $i=0$ \textbf{ then }\\
		\hspace{0.5in}Select controller based on equation (\ref{eq:OptimalControlExperiment})
		\STATE\textbf{else}\\
		\hspace{0.5in}Run policy $\pi_\theta$
		\STATE\textbf{end}
	\end{algorithmic}
\end{algorithm}

This action selection method ensures that we select actions that are not only in $\{-1,1\}$ but also in $[-1,1]$. It introduces a new method of exploration of the environment by choosing from two different methods of action selection. Actions selected using equation \ref{eq:OptimalControlExperiment} are also stored in the memory buffer and are used to train the policy network $\pi_\theta$. The method is provided in Algorithm \ref{alg:HJBPPO}.

\begin{algorithm}[tb]
	\caption{HJBPPO}
	\label{alg:HJBPPO}
	\begin{algorithmic}[1]
		\STATE Initiate policy network parameter $\theta$ and value network parameter $\phi$
		\STATE Run action selection as given in algorithm \ref{alg:action} in the environment for $T$ timesteps and observe samples $\{(s_t,a_t,R_t,s_{t+1})\}_{t=1}^{T}$.
		\STATE Compute the advantage $A_t$
		\STATE Compute $r_t(\theta)=\frac{\pi_\theta(a_t|s_t)}{\pi_{\theta_{\text{old}}}(a_t|s_t)}$
		\STATE Compute the objective function of the policy network:
		\begin{align*}
			L(\theta)=\frac{1}{T}\sum_{t=0}^{T-1}&\min[r_t(\theta)A_t,\text{clip}(r_t(\theta),1-\epsilon,1+\epsilon)A_t],
		\end{align*}
		\STATE Update $\theta\leftarrow\theta+\alpha_1\nabla_\theta L(\theta)$
		\STATE Compute the value network loss as: $J(\phi)=MSE_f+MSE_u+MSE_n$ described in equations (\ref{eq:HJBLoss2}), (\ref{eq:HJBInitLoss}), and (\ref{eq:HJBNeumannLoss})
		\STATE Update $\phi\leftarrow\phi-\alpha_2\nabla_\phi J(\phi)$
		\STATE Run steps 2--8 for multiple iterations
	\end{algorithmic}
\end{algorithm}

We will train PPO, HJB value iteration, and HJBPPO on the PackCooling environment and compare these algorithms.

\section{Results}

\subsection{Training}

To ensure the reproducibility of our results, we have posted our code in the following link: 

\noindent\href{https://github.com/amartyamukherjee/PPO-PackCooling}{https://github.com/amartyamukherjee/PPO-PackCooling}. We posted our hyperparameters in Appendix \ref{app:hparams}. The details of the implementation of the PackCooling gym environment are posted in Appendix \ref{app:gym}. The code was run using Kaggle CPUs. Each algorithm was trained for a million timesteps. Training each algorithm took approximately 5 hours.

\subsection{Reward Curves}

\begin{figure}[tb]
	\centering
	\includegraphics[width=0.9\linewidth]{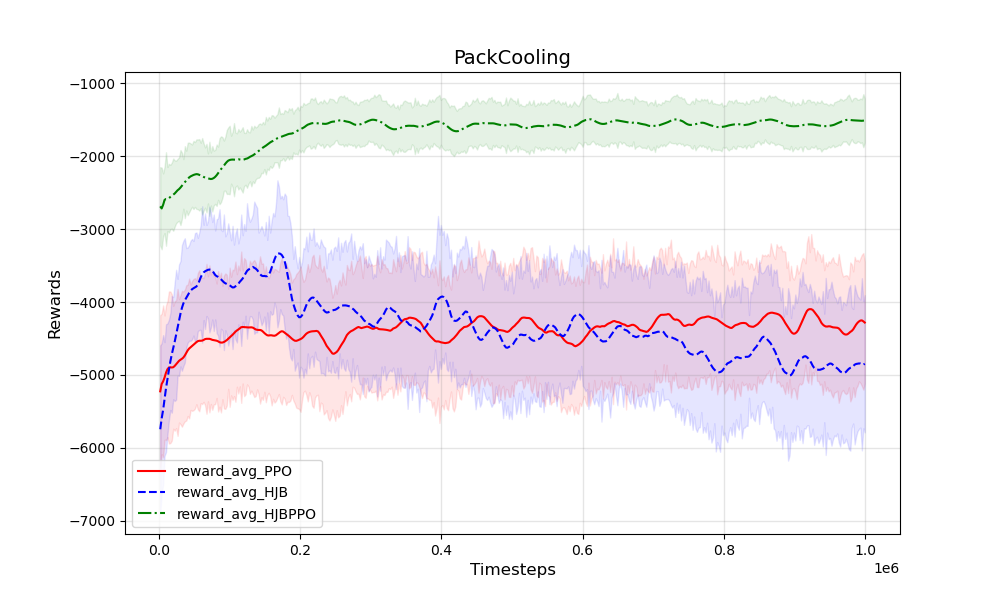}
	\caption{Reward curves of PPO (red), HJB value iteration (blue), and HJBPPO (green) averaged over 5 seeds. Shaded area indicates 0.2 standard deviations.}
	\label{fig:reward}
\end{figure}

The reward curves have been plotted in Figure \ref{fig:reward}, comparing PPO, HJB value iteration, and HJBPPO. Each algorithm was run for 5 different seeds. We plotted the mean reward over each seed and over 20 consecutive episodes, and shaded the area 0.2 standard deviations from the mean. HJB value iteration shows the worst performance, as its rewards decrease past PPO after training for multiple episodes. PPO shows a rapid increase in average rewards after the first episode and a slow increase in average rewards afterward. HJBPPO shows the best performance in the graph, achieving the highest average reward in each episode and an increase in average rewards after training for multiple episodes.

The significantly higher average reward in HJBPPO in the first episode shows that the action selection method described in Algorithm \ref{alg:action} provides a robust strategy to explore the environment and train the models. The higher average rewards are due to the exploitation of the dynamics of the environment as done by the HJB equation.

\subsection{Trajectories}

The plots of the trajectories have been posted in Appendix \ref{app:trajectories}. After training for a million timesteps, we tested our models on the PackCooling environment and produced the plots. These plots were generated using the rendering feature explained in section \ref{sec:rendering}.

The trajectory of HJB value iteration shows the worst results. $\sigma(t)$ returns $1.0$ only once. It achieves a cumulative reward of $-7294.51$. Thus, the input of the cooling fluid from the boundary is minimal. As a result of the internal heat generation in the battery pack, $u(x,t)$ reaches high values of roughly $5$ at $t=10$, and as a result, $w(x,t)$ also reaches high values of roughly $4$. This shows that the training of the value function in HJB value iteration is inadequate and we have not arrived at an optimal controller for the pack cooling problem. This is because exploration of the environment was at a minimum, as we only exploited equation \ref{eq:OptimalControlExperiment} at each time step.

The trajectory of PPO shows that the values $\sigma(t)$ takes at every timestep have a large variance with its neighboring timesteps. It achieves a cumulative reward of $-3970.02$. Control of the temperature of the battery pack has been achieved as $u(x,t)$ takes values between $-2$ and $2$ at $t=10$.

The trajectory of u(x,t) with HJBPPO shows it takes values between $-2$ and $2$ at $t=10$. The values $\sigma(t)$ takes at every timestep have a lower variance with its neighboring timesteps compared to PPO. It achieves a cumulative reward of $-881.55$. For $t\in[4,6]$, $u(x,t)$ shows an increasing trend towards $u=2$. In response, the controller $\sigma(t)$ took values closer to $1.0$ to allow for greater input of cooling fluid from the boundary so that $u(x,t)$ decreases towards zero. Due to higher average rewards as shown in Figure \ref{fig:reward}, this shows that a model that exploits the dynamics of the environment to return a controller shows improved performance compared to a model that returns noisy control centered at $\sigma=0.5$.

\section{Conclusion}

In this paper, we have introduced two algorithms that use PINNs to solve the pack cooling problem. This paper combines PINNs with RL in a PDE control setting. In the HJB value iteration algorithm, the HJB equation is used to introduce a loss function and a controller using a value network. The HJBPPO algorithm is a hybrid-policy model that combines the training of the value network from HJB value iteration and the training of the policy network from PPO. HJBPPO shows an overall improvement in performance compared to PPO due to its ability to exploit the physics of the environment to improve the learning curve of the agent.

\section{Future research}

Despite showing an overall improvement in the reward curves, the HJBPPO algorithm leaves room for improved RL algorithms using PINNs.

In this paper, we computed the HJB equation by expressing the PDE as an ODE by discretizing in $x$. This was possible because the pack cooling problem was modeled by 1D PDEs. Currently existing works such as (\cite{SIRIGNANO20181339}) and (\cite{HJBPDE}) solve the HJB equation for 1D PDEs by discretizing it in $x$. It will be interesting to see how HJB control can be extended to higher dimensional PDEs.

The goal of PINNs is to solve PDEs without the need for numerical methods. In this paper, we solved the pack cooling problem numerically using the Crank-Nicolson method and the method of characteristics. An area for further research may be the use of PINNs to solve for the HJB equation and the PDE that governs the dynamics of the system.

In the PackCooling environment, the HJBPPO algorithm showed an improvement compared to PPO. But this is due to the fact that we knew the dynamics of the system, thus allowing for the physics of the environment to be exploited. The environments give all the details of the state needed to choose an action. One limitation of HJBPPO is that it may not perform well in partially observable environments because the estimate of the dynamics of the system may be inaccurate. Deep Transformer Q Network (DTQN) was introduced by \cite{DTQN} and achieves state-of-the-art results in many partially observable environments. A potential area for further research may be the introduction of an HJB equation that facilitates partial observability. The DTQN algorithm may be improvised by incorporating this HJB equation using PINNs.

\bibliography{example_paper}
\bibliographystyle{icml2023}

%%%%%%%%%%%%%%%%%%%%%%%%%%%%%%%%%%%%%%%%%%%%%%%%%%%%%%%%%%%%%%%%%%%%%%%%%%%%%%%
%%%%%%%%%%%%%%%%%%%%%%%%%%%%%%%%%%%%%%%%%%%%%%%%%%%%%%%%%%%%%%%%%%%%%%%%%%%%%%%
% APPENDIX
%%%%%%%%%%%%%%%%%%%%%%%%%%%%%%%%%%%%%%%%%%%%%%%%%%%%%%%%%%%%%%%%%%%%%%%%%%%%%%%
%%%%%%%%%%%%%%%%%%%%%%%%%%%%%%%%%%%%%%%%%%%%%%%%%%%%%%%%%%%%%%%%%%%%%%%%%%%%%%%
\newpage
\appendix
\onecolumn

\section{The PackCooling gym environment}
\label{app:gym}

To run numerical experiments in this paper, we had to create a gym environment that solves the pack cooling model numerically and takes a controller as input every time step.

In gym environments, it is recommended to set the action range to $[-1,1]$ (\cite{rl_tips}). In our control problem, $\sigma(t)$ returns values in $[0,1]$. Thus, we let $\sigma(t)=\frac{a_t+1}{2}$.

\subsection{Internal heat generation}

In this paper, we set the internal heat generation in the battery pack, $h(x,t,u(x,t))$ as:

\begin{equation}
	h(x,t,u(x,t))=e^{0.1u(x,t)}. 
\end{equation}

The objective of the internal heat generation was to increase the temperature distribution across the battery pack at every time step, which serves as a motivation for choosing a strictly positive exponential function. This makes the desired state of the controller, $u=0$, unstable.

\subsection{Initial conditions}

For $u(x,t)$ we chose initial conditions that satisfy the boundary conditions \ref{eq:bc1}. A set of solutions that satisfy the boundary conditions are:
\begin{equation}
	u(x,0)=\sum_{n=0}^{\infty}C_n\cos(\pi nx),
\end{equation}
since the derivative of $\cos(\pi nx)$ is zero at $x=0$ and $x=1$ for all integers $n$. Thus, we set our initial conditions to the following:
\begin{equation}
	u(x,0)=\sum_{n=0}^{N}C_n\cos(\pi nx), 
\end{equation}
where $N$ is the number of Fourier nodes. Note that $N\leq N_x$ where $N_x$ is the number of points we discretize the numerical solution into. In our experiments, we set $N=9$. The Fourier coefficients $C_n$ are sampled from the uniform distribution $U(-2,2)$. Adding randomness in our initial conditions leads to more robustness in our controller.

For $w(x,t)$, we set the initial condition to $w(x,0)=U(0)$ where $U$ is defined in the boundary condition in equation \ref{eq:bc2}. In our experiments, we set $U(t)$ to a constant: $U(t)=-5.0$.

This means if we set the controller to a constant $\sigma(t)=0$, then $u(x,t)$ will increase to high values due to the forcing term $h(x,t,u(x,t))$. And if we set the controller to a constant $\sigma(t)=1$, then $u(x,t)$ will decrease to low values due to the cooling fluid. So the objective of the controller is to choose $\sigma(t)$ so that the temperature distribution in the battery pack is as close to zero as possible.

\subsection{Numerical solution}

In our numerical solutions, we discretize the heat distributions $u(x,t)$ and $w(x,t)$ in $x$ and $t$ as: $\Delta x=0.01,\Delta t=0.01$, thus satisfying $\Delta x\geq \Delta t\geq\sigma(t)\Delta t$. We set $D(x,t)$ and $R(x,t)$ from equations \ref{eq:PDE1} and \ref{eq:PDE2} to constants as $D(x,t)=0.01, R(x,t)=2.0$.

The numerical solution to the PDE was implemented as explained in \cite{Kato_2021}. We solve for a system 

\begin{equation}
	A^+z(t+\Delta t)-\frac{1}{2}h(z(t+\Delta t))=Az(t)+\frac{1}{2}h(z(t)),
\end{equation}
where $z(t)$ is the concatenation of $u(x,t)$ and $w(x,t)$ discretized in space. The matrices $A^+$ and $A$ are derived using a combination of the Crank-Nicolson method, characteristics, and interpolation. Finally, to solve this non-linear system, we run $10$ iterations of the Newton-Raphson method.

\subsection{Rendering}
\label{sec:rendering}

In the rendering of the environment, we plot the trajectories of $\sigma(t),u(x,t),w(x,t)$ in the environment. We store the numerical solutions of the three variables in a buffer every episode and reset the buffer upon resetting the environment.

Examples of rendered plots are shown in Appendix \ref{app:trajectories}. We return a line plot of $\sigma(t)$ along $t$, and a mesh plot of $u(x,t)$ and $w(x,t)$ along $x$ and $t$.

\section{Proof of theorem \ref{thm:1} part 1}
\label{app:pf1}

\textbf{Theorem}
\textit{A function $V(x)$ is the optimal value function if and only if $V\in C^1(\mathbb{R}^n)$ and $V$ satisfies the HJB Equation}
\begin{align}
    &(\gamma-1)V(x)+\sup_{\sigma\in U}\{L(x,\sigma)+\gamma\Delta t\nabla_xV^T(x)f(x,\sigma)\}=0
\end{align}
\textit{for all $x\in\mathbb{R}^n$, assuming for all $x\in\mathbb{R}^n$, there exists a controller $\sigma^*(\cdot)$ such that:}
\begin{align}
    &(\gamma-1)V(x)+L(x,\sigma^*(x))+\gamma\Delta t\nabla_xV^T(x)f(x,\sigma^*(x))
    \nonumber\\&=(\gamma-1)V(x)+\sup_{\hat \sigma\in U}\{L(x,\hat \sigma)+\gamma\Delta t\nabla_xV^T(x)f(x,\hat\sigma)\}.
\end{align}

\textbf{Proof} We start with equation \ref{eq:value1}:

\begin{equation}
	V^*(x(t_0))=\sup_{\sigma}\frac{1}{\Delta t}\int_{t_0}^{\infty}\gamma^{\frac{t}{\Delta t}} L(x(\tau),\sigma(\tau))d\tau
\end{equation}

Separating the integral term gives:

\begin{align}
	V^*(x(t_0))&=\sup_{\sigma}\left[\frac{1}{\Delta t}\int_{t_0}^{t_0+\Delta t}\gamma^{\frac{t}{\Delta t}} L(x(\tau),\sigma(\tau))d\tau+\frac{\gamma}{\Delta t}\int_{t_0+\Delta t}^{\infty}\gamma^{\frac{t}{\Delta t}} L(x(\tau),\sigma(\tau))d\tau\right]\\
    &=\sup_{\sigma}\left[\frac{1}{\Delta t}\int_{t_0}^{t_0+\Delta t}\gamma^{\frac{t}{\Delta t}} L(x(\tau),\sigma(\tau))d\tau+\gamma V^*(x(t_0+\Delta t))\right]
\end{align}

As $\Delta t\to 0$, we can rely on the following two assumptions:

\begin{equation}
    \frac{1}{\Delta t}\int_{t_0}^{t_0+\Delta t}\gamma^{\frac{t}{\Delta t}} L(x(\tau),\sigma(\tau))d\tau\approx\frac{1}{\Delta t}(\Delta tL(x(t_0),\sigma(t_0)))=L(x(t_0),\sigma(t_0))
\end{equation}
\begin{equation}
    V^*(x(t_0+\Delta t))\approx V^*(x(t_0))+\Delta t\nabla_xV^*(x(t_0))^Tf(x(t_0))
\end{equation}

This gives us the following equation:

\begin{equation}
    V^*(x(t_0))=\sup_{\sigma}\left[L(x(t_0),\sigma(t_0))+\gamma V^*(x(t_0))+\gamma\Delta t\nabla_xV^*(x(t_0))^Tf(x(t_0))\right]
\end{equation}
\begin{equation}
    0=(\gamma-1)V^*(x(t_0))+\sup_{\sigma}\left[L(x(t_0),\sigma(t_0))+\gamma\Delta t\nabla_xV^*(x(t_0))^Tf(x(t_0))\right]
\end{equation}

\section{Proof of theorem \ref{thm:HJB}}
\label{app:pfThm}

\textbf{Theorem}
	\textit{Let $u(\cdot,t),w(\cdot,t)\in L_2[0,1]$. With $\sigma(t)\in[0,1]$ and the reward function $L(U_t,W_t,\sigma_t)=-||U_{t+1}||_2^2\Delta x$, the HJB equation for the 1D pack cooling problem is:}
	\begin{align}
		&(1-\gamma)V(u,w)-||u(\cdot,t+\Delta t)||^2+\langle V_u,u_t\rangle+\frac{1}{R}\langle V_w,u-w\rangle+\max(0,-\langle V_w,w_x\rangle)=0
	\end{align}
	\textit{where $||\cdot||$ is the $L_2[0,1]$ norm and $\langle\cdot,\cdot\rangle$ is the $L_2[0,1]$ inner product.}

\textbf{Proof} In the pack cooling problem, the value function is a function of $\hat{U}(t)$ and $\hat{W}(t)$, which is $u(x,t)$ and $w(x,t)$ discretized in $x$. The HJB equation from equation \ref{eq:HJB1} can thus be written as follows:

\begin{equation}
    (\gamma-1)V(\hat{U}(t),\hat{W}(t))+\sup_{\sigma\in[0,1]}\{L(\hat{U}(t),\hat{W}(t),\sigma)+\gamma\Delta t\nabla_{\hat{U}}V^T(\hat{U}(t),\hat{W}(t))\dot{\hat{U}}+\gamma\Delta t\nabla_{\hat{W}}V^T(\hat{U}(t),\hat{W}(t))\dot{\hat{W}}\}=0
\end{equation}

We know that $L(\hat{U}(t),\hat{W}(t),\sigma)=-||\hat{U}(t+\Delta t)||_2^2\Delta x$, where $\Delta x$ and $\Delta t$ are the discretizations in $x$ and $t$ used in the environment respectively. $\hat{U}(t+\Delta t)$ is determined by $\hat{U}(t)$ and $\dot{\hat{U}}(t)$. We know from equation \ref{eq:ODE1} that neither $\hat{U}(t)$ or $\dot{\hat{U}}(t)$ depend on $\sigma(t)$. Thus, we can bring $L(\hat{U}(t),\hat{W}(t),\sigma)$ and $\nabla_{\hat{U}}V^T(\hat{U}(t),\hat{W}(t))\dot{\hat{U}}$ outside the supremum.

\begin{equation}
    (\gamma-1)V(\hat{U}(t),\hat{W}(t))+L(\hat{U}(t),\hat{W}(t),\sigma)+\gamma\Delta t\nabla_{\hat{U}}V^T(\hat{U}(t),\hat{W}(t))\dot{\hat{U}}+\sup_{\sigma\in[0,1]}\{\gamma\Delta t\nabla_{\hat{W}}V^T(\hat{U}(t),\hat{W}(t))\dot{\hat{W}}\}=0
\end{equation}

We expand $\dot{\hat{W}}$ from equation \ref{eq:ODE2}.

\begin{align}
	\sup_{\sigma\in [0,1]}\{\nabla_{\hat{W}}V^T(\hat{U},\hat{W})\dot{\hat{W}}\}
	&=\sup_{\sigma\in [0,1]}\{\nabla_{\hat{W}}V^T(\hat{U},\hat{W})(-\sigma(t)B\hat W+\frac{1}{R}(\hat U-\hat W))\}\\
	&=\nabla_{\hat{W}}V^T(\hat{U},\hat{W})(\frac{1}{R}(\hat U-\hat W))+\sup_{\sigma\in [0,1]}\{\nabla_{\hat{W}}V^T(\hat{U},\hat{W})(-\sigma(t)B\hat W)\}
\end{align}

For the expression to achieve its supremum, we let $\sigma(t)$ be defined as the following:

\begin{equation}
	\label{eq:appOptimalControl}
	\sigma(t)=\begin{cases}1&\nabla_{\hat{W}}V^T(\hat{U},\hat{W})B\hat W<0\\0&\text{otherwise}\end{cases}
\end{equation}

We get:
\begin{equation}
	\sup_{\sigma\in [0,1]}\{\nabla_{\hat{W}}V^T(\hat{U},\hat{W})(-\sigma(t)B\hat W)\}=\max(0,-\nabla_{\hat{W}}V^T(\hat{U},\hat{W})B\hat W)
\end{equation}

Thus, we get the following HJB equation in discretized $x$:

\begin{equation}
    \label{eq:HJBexperiment}
    (\gamma-1)V(\hat{U},\hat{W})+L(\hat{U},\hat{W},\sigma)+\gamma\Delta t\nabla_{\hat{U}}V^T(\hat{U},\hat{W})\dot{\hat{U}}+\nabla_{\hat{W}}V^T(\hat{U},\hat{W})(\frac{1}{R}(\hat U-\hat W))+\max(0,-\gamma\Delta t\nabla_{\hat{W}}V^T(\hat{U},\hat{W})B\hat W)=0
\end{equation}

Let $\Delta x=\Delta t$ to satisfy the condition for numerical stability. As we let $\Delta x\to 0$, we see a Riemann sum forming.

\begin{equation}
	(\gamma-1)V-\int_0^1(u(\cdot,t+\Delta t))^2dx+\int_0^1V_uu_tdx+\int_0^1V_w(\frac{1}{R}(u-w))dx+\max(0,-\int_0^1V_ww_xdx)=0
\end{equation}

We can write this as inner products and norms in the $L_2[0,1]$ space.

\begin{equation}
	(\gamma-1)V-||u(\cdot,t+\Delta t)||^2+\langle V_u,u_t\rangle+\langle V_w,\frac{1}{R}(u-w)\rangle+\max(0,-\langle V_w,w_x\rangle)=0
\end{equation}

\section{Proof of corollary \ref{cor:OptimalController}}
\label{app:pfCor}

\textbf{Corollary}
	\textit{Let $u(\cdot,t),w(\cdot,t)\in L_2[0,1]$. With $\sigma(t)\in[0,1]$ and the reward function $L(U_t,W_t,\sigma_t)=-||U_{t+1}||_2^2\Delta x$, provided the optimal value function $V^*(u,w)$ the optimal controller for the 1D pack cooling problem is:}
	\begin{equation}
		\sigma^*(t)=\begin{cases}1,&\langle V^*_w(\cdot,t),w_x(\cdot,t)\rangle<0,\\0,&\text{otherwise},\end{cases}
	\end{equation}
	\textit{where $||\cdot||$ is the $L_2[0,1]$ norm and $\langle\cdot,\cdot\rangle$ is the $L_2[0,1]$ inner product.}

\textbf{Proof}
The optimal controller was derived in the proof of the HJB equation in equation \ref{eq:appOptimalControl}. As we let $\Delta x\to 0$, we get:

\begin{equation}
	\sigma(t)=\begin{cases}1&\int_0^1V_ww_xdx<0\\0&\text{otherwise}\end{cases}
\end{equation}

We can write this as inner products and norms in the $L_2[0,1]$ space.

\begin{equation}
	\sigma(t)=\begin{cases}1&\langle V_w,w_x\rangle<0\\0&\text{otherwise}\end{cases}
\end{equation}

\section{Hyperparameters}
\label{app:hparams}

\begin{table}[H]
	\centering
	\begin{tabular}{ c|c } 
		Hyperparameter & Value\\
		\hline
		Horizon (T) & 1024 \\ 
		Actor learning rate & 3e-04 \\ 
		Critic learning rate & 1e-03 \\ 
		Num.  epochs & 10 \\ 
		Minibatch size & 64 \\ 
		Discount ($\gamma$) & 0.99 \\ 
		GAE parameter ($\lambda$) & 0.95 \\ 
	\end{tabular}
	\caption{Hyperparameters}
	\label{tab:hparams}
\end{table}

\section{Plots of trajectories}
\label{app:trajectories}

\begin{figure}[H]
	\centering
	\includegraphics[height=0.7\paperheight]{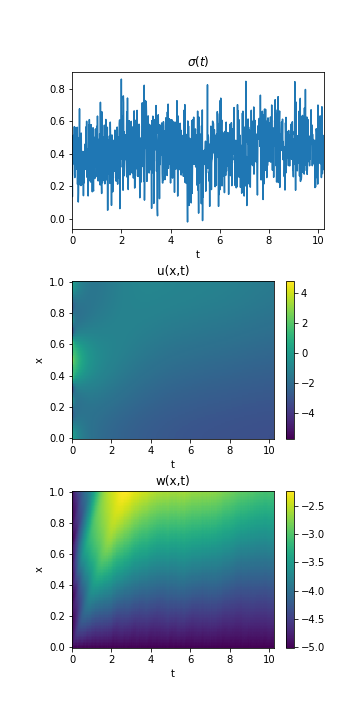}
	\caption{Trajectory of PPO. Cumulative reward: $-3970.02$}
	\label{fig:trajectory_ppo}
\end{figure}

\begin{figure}[H]
	\centering
	\includegraphics[height=0.7\paperheight]{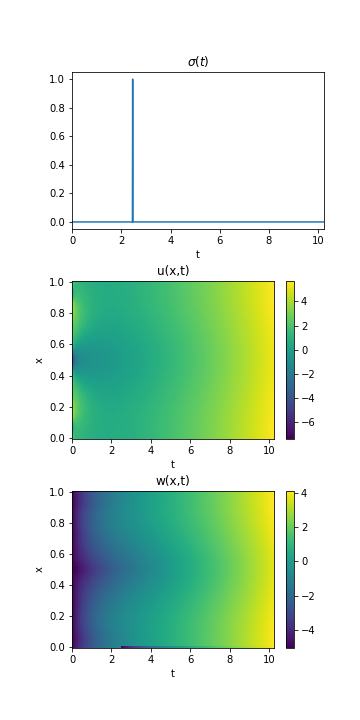}
	\caption{Trajectory of HJB value iteration. Cumulative reward: $-7294.51$}
	\label{fig:trajectory_hjb}
\end{figure}

\begin{figure}[H]
	\centering
	\includegraphics[height=0.7\paperheight]{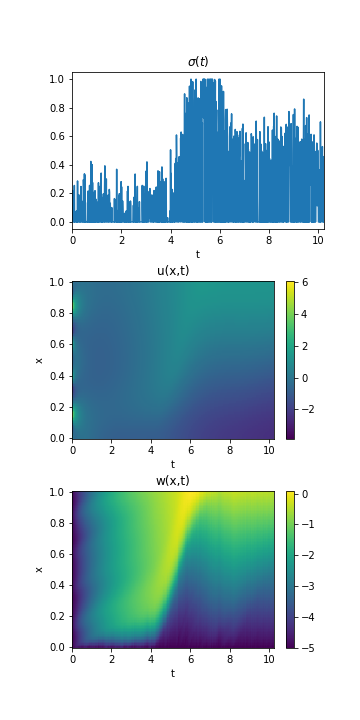}
	\caption{Trajectory of HJBPPO. Cummulative reward: $-881.55$}
	\label{fig:trajectory_hjbppo}
\end{figure}

%%%%%%%%%%%%%%%%%%%%%%%%%%%%%%%%%%%%%%%%%%%%%%%%%%%%%%%%%%%%%%%%%%%%%%%%%%%%%%%
%%%%%%%%%%%%%%%%%%%%%%%%%%%%%%%%%%%%%%%%%%%%%%%%%%%%%%%%%%%%%%%%%%%%%%%%%%%%%%%

\end{document}